\documentclass{esannV2}
\usepackage[dvips]{graphicx}
\usepackage[latin1]{inputenc}
\usepackage{amssymb,amsmath,array}
\usepackage{booktabs}
\usepackage{multirow}
\usepackage{subcaption}
\usepackage{url}

\newcommand{\meanstd}[2]{${#1}_{\pm #2}$}
\newcommand{\boldmeanstd}[2]{$\mathbf{{#1}_{\pm #2}}$}

\begin{document}
\title{Continual Learning with Echo State Networks}

\author{Andrea Cossu$^{1,2}$, Davide Bacciu${^1}$, Antonio Carta$^1$,\\Claudio Gallicchio${^1}$ and Vincenzo Lomonaco${^1}$

\thanks{This work has been partially supported by the H2020 TEACHING project (GA 871385).}

\vspace{.3cm}\\
1- University of Pisa - Department of Computer Science \\
Largo B. Pontecorvo, 3, 56127, Pisa - Italy
\vspace{.1cm}\\
2- Scuola Normale Superiore \\
Piazza dei Cavalieri, 7, 56126, Pisa - Italy \\
}
\maketitle

\begin{abstract}
Continual Learning (CL) refers to a learning setup where data is non stationary and the model has to learn without forgetting existing knowledge. The study of CL for sequential patterns revolves around trained recurrent networks. In this work, instead, we introduce CL in the context of Echo State Networks (ESNs), where the recurrent component is kept fixed.
We provide the first evaluation of catastrophic forgetting in ESNs and we highlight the benefits in using CL strategies which are not applicable to trained recurrent models. Our results confirm the ESN as a promising model for CL and open to its use in streaming scenarios.
\end{abstract}


\section{Introduction}
Real world environments where data is highly non stationary represent a challenge for current machine learning solutions. Continual Learning (CL) focuses on the design of new models and techniques able to learn new information while preserving existing knowledge \cite{lesort2020}. CL models receive a (possibly infinite) stream of experiences $e_1, e_2, e_3, ...$, where each experience $e_i$ contains data sampled from an associated distribution $D_i$. Drifts in data distribution usually occur during a known transition to a new experience. We study CL from the perspective of catastrophic forgetting: when trained sequentially on multiple experiences, neural networks tend to forget previous knowledge, that is, they reduce their performance on previously seen data and tasks. \\
We address CL in sequential data processing (where each pattern is a sequence) with a family of randomized recurrent neural networks called Echo State Networks (ESNs) \cite{Jaeger2004}. ESNs process sequences by means of a fixed recurrent component initialized with stable dynamics (reservoir) whose output is fed to a trainable output layer (readout). ESNs are promising architectures for challenging contexts such as neuromorphic hardware and embedded intelligence applications. \\
The topic of CL with recurrent neural networks has started to gain attention from the CL community. Recent works on the behavior of recurrent models in non stationary environments highlighted the fact that common CL strategies may behave differently than expected when applied to recurrent models (e.g. their performance may be influenced by the sequence length) \cite{cossu2021}. While these studies focus on fully trained recurrent models, we instead provide the first experimental analysis of catastrophic forgetting in ESNs, and their use with popular CL strategies. By treating the untrained reservoir of an ESN as a feature extractor for sequences, we were able to 1) restrict the application of CL strategies to the final, linear readout and 2) to leverage efficient CL strategies operating solely on the final layer. The latter point marks a clear advantage in using ESNs, since many approaches in CL for computer vision often exploit pretrained feature extractors. While such methods could be directly applied to ESNs, it is generally difficult to do the same for other recurrent models, where the use of pretrained networks is not as effective. Our results indicate that ESN exhibits competitive performance with respect to LSTM and it is amenable to be applied with CL solutions operating in challenging settings like streaming learning. 

\section{Continual Learning and Recurrent Models}
\paragraph{Continual Learning} The study of recurrent models in CL currently focuses on deep recurrent networks trained by backpropagation \cite{cossu2021, sodhani2019}. The problem has been tackled both by designing new techniques and learning algorithms \cite{duncker2020} and by applying popular CL strategies not designed for sequential data \cite{cossu2020, sodhani2019}. Notably, the CL performance of alternative recurrent paradigms like spiking neural networks \cite{ororbiaSpikingNeuralPredictive2020} and reservoir computing \cite{kobayashi2019} is under-documented. To the best of our knowledge, there is only one work about CL and ESNs \cite{kobayashi2019} which, however, focuses on a specific application and does not give insights on how to use ESNs in different CL contexts. This highlights the need for a broad experimental evaluation with different families of CL strategies on popular benchmarks.

\paragraph{Continual Learning strategies}
Here, we briefly introduce the CL strategies used in our experiments. These strategies are not specifically designed for sequential data processing. Therefore, they allow us to draw more general conclusions on the model behavior.\\
Elastic Weight Consolidation (EWC) \cite{kirkpatrick2017} and Learning without Forgetting (LwF) \cite{li2016} are regularization strategies: they add a penalty to the loss function to improve model stability. EWC prevents large changes in parameters deemed important for previous experiences. The loss penalty at experience $n$ takes the form of $\sum_{t=1}^{n-1} \mathbf{\Omega_t} (\mathbf{\theta_t} - \mathbf{\theta_n})^2,$ where $\mathbf{\theta_t}$ are the model parameters at experience $t$ and $\mathbf{\Omega}$ their corresponding importances. At the end of each experience, the parameters importances for that experience are computed by freezing the model, performing an additional pass over training data and averaging the squared gradients across all patterns. \\
LwF enforces stability in the output layer activations by keeping a copy of the previous model and by taking its output on the current training patterns as soft targets in the distillation loss. Therefore, the loss penalty takes the form of the the KL-divergence $\text{KL}[p_\mathbf{\theta_t}(\mathbf{x_t}) || p_\mathbf{\theta_{t-1}}(\mathbf{x_t})],$ where $p_\mathbf{\theta_t}(\mathbf{x_t})$ represents the output of the model parameterized by $\mathbf{\theta_t}$.\\
Replay \cite{robins1995} leverages a different paradigm: on each experience, it randomly samples a number of patterns from the training set and adds them to the replay memory. During training, the current minibatch is augmented with an additional minibatch of patterns sampled directly from the memory. This is one of the most effective strategies in CL \cite{prabhu2020}. \\
Finally, Streaming Linear Discriminant Analysis (SLDA) \cite{hayes2020} is a strategy which leverages a pretrained feature extractor $G$ (a ResNet-18 in the original paper) combined with a linear layer to compute the final output $\mathbf{y} = \mathbf{W} G(\mathbf{x}) + \mathbf{b}$, where $\mathbf{x}$ is the current input pattern. This strategy operates in a streaming setting, where patterns are seen one at a time and only once (one epoch only). The parameters $\mathbf{W}$ and $\mathbf{b}$ of the linear layer are computed through an online approximation of the Linear Discriminant Analysis. The algorithm keeps a mean vector together with an associated counter per class and a shared covariance matrix, updated during training. Since SLDA requires a fixed feature extractor, we can only apply it to ESNs thanks to its untrained recurrent layer. \\
To provide lower and upper bounds performance, we ran experiments with other $2$ strategies: Naive, which simply finetunes the network across experiences without any CL strategy, and Joint Training, which trains the model in an offline setting with all the data available from the beginning. 

\paragraph{Echo State Networks} Reservoir Computing provides a general framework to build recurrent networks \cite{lukoseviciusReservoirComputingApproaches2009}. ESNs \cite{Jaeger2004} belong to the reservoir computing paradigm since they are composed by an untrained reservoir and a trained linear readout. The reservoir is composed by a set of randomly connected units and it represents the recurrent component of the architecture. The initialization of recurrent weights matrix in the reservoir is a crucial hyperparameter: usually, the matrix values are scaled such that its spectral radius is slightly smaller than one (necessary condition for the Echo State Property). The linear readout parameters can be trained with closed-form solutions like pseudo-inverse or ridge regression, which however requires to store the entire set of reservoir activations.

\section{Experiments}
We compared the performance of LSTM and ESN when equipped with the $4$ CL strategies already presented: EWC, LwF, Replay, SLDA. We tested our methods on $2$ different sequence classification tasks, in which each pattern (a sequence) is associated to a target class. We chose $2$ popular class-incremental CL benchmarks, where each experience provides examples from new classes, which will be present only in that experience. At the end of training on the last experience, the model performance is measured against data coming from all experiences. During testing, the model has no knowledge about the experience from which each pattern is coming from. 

\paragraph{Experimental setup}
We used Split MNIST (SMNIST) and SSC \cite{cossu2021} as the benchmarks for our experiments. SMNIST provides $5$ different experiences, each of which contains examples of MNIST dataset from $2$ digit classes. In order to use SMNIST as a sequence classification task, we took each image one row at a time, resulting in input sequences with $28$ steps. SSC is a dataset of spoken words. We took $10$ experiences, each of which containing patterns representing $2$ words. Sequences have $101$ steps. We performed grid search on all the strategies for ESN and on Replay for LSTM. The other results for LSTM are taken from \cite{cossu2021}, since the experimental setup is the same. To perform grid search for SSC, we took $3$ held-out experiences for model selection and $10$ for model assessment. To fairly compare ESN and LSTM, we select a model configuration whose only requirement is to be able to learn effectively at training time. Then, the performance in terms of forgetting depends mostly on the CL strategy (subjected to grid search) and not on the specific model setting. We train the ESN readout with Adam optimizer and backpropagation, since CL requires to update the model continuously, possibly without storing its activations. We used the Avalanche \cite{lomonaco2021} framework for all our CL experiments. We make publicly available the code together with configurations needed to reproduce all experiments\footnote{\url{https://github.com/Pervasive-AI-Lab/ContinualLearning-EchoStateNetworks}}. We monitored the average accuracy (ACC) metric: after training on all experiences we measure the accuracy averaged over test patterns from all the experiences.

\paragraph{Results}
\begin{table}[t]
\centering
\begin{tabular}{lcc}
     \toprule
     \textbf{SMNIST} & LSTM$^\dagger$ & ESN  \\
     \midrule
     EWC    & \meanstd{0.21}{0.02} & \meanstd{0.20}{0.00} \\
     LWF    & \meanstd{0.31}{0.07} & \meanstd{0.47}{0.07} \\
     REPLAY & \boldmeanstd{0.85}{0.03} & \meanstd{0.74}{0.03} \\
     SLDA  & --- & \boldmeanstd{0.88}{0.01}\\ 
     \midrule
     NAIVE & \meanstd{0.20}{0.00} & \meanstd{0.20}{0.00}\\
     JOINT & \meanstd{0.97}{0.00} & \meanstd{0.97}{0.01} \\
     \bottomrule
\end{tabular}
\hfill
\begin{tabular}{lcc}
    \toprule
    \textbf{SSC} & LSTM$^\dagger$ & ESN  \\
    \midrule
     EWC & \meanstd{0.10}{0.00} & \meanstd{0.09}{0.02} \\
     LWF & \meanstd{0.12}{0.01} & \meanstd{0.12}{0.02} \\
     REPLAY & \boldmeanstd{0.74}{0.07} & \meanstd{0.36}{0.07} \\
    SLDA  & --- & \boldmeanstd{0.57}{0.03} \\ \midrule
     NAIVE & \meanstd{0.10}{0.00} & \meanstd{0.10}{0.00} \\
     JOINT & \meanstd{0.89}{0.02} & \meanstd{0.91}{0.02}  \\   
     \bottomrule
\end{tabular}
\caption{Mean ACC and standard deviation over $5$ runs on SMNIST and SSC benchmarks. SLDA is applied only to ESN since it assumes a fixed feature extractor. SMNIST contains $5$ experiences, while SSC contains $10$ experiences. $\dagger$ results are taken from \cite{cossu2021}, except for replay which has been recomputed to guarantee the use of the same replay policy ($200$ patterns in memory).}
\label{tab:results}
\end{table}

\begin{figure}[t]
\centering
\begin{subfigure}{.42\textwidth}
    \centering
    \includegraphics[width=\textwidth]{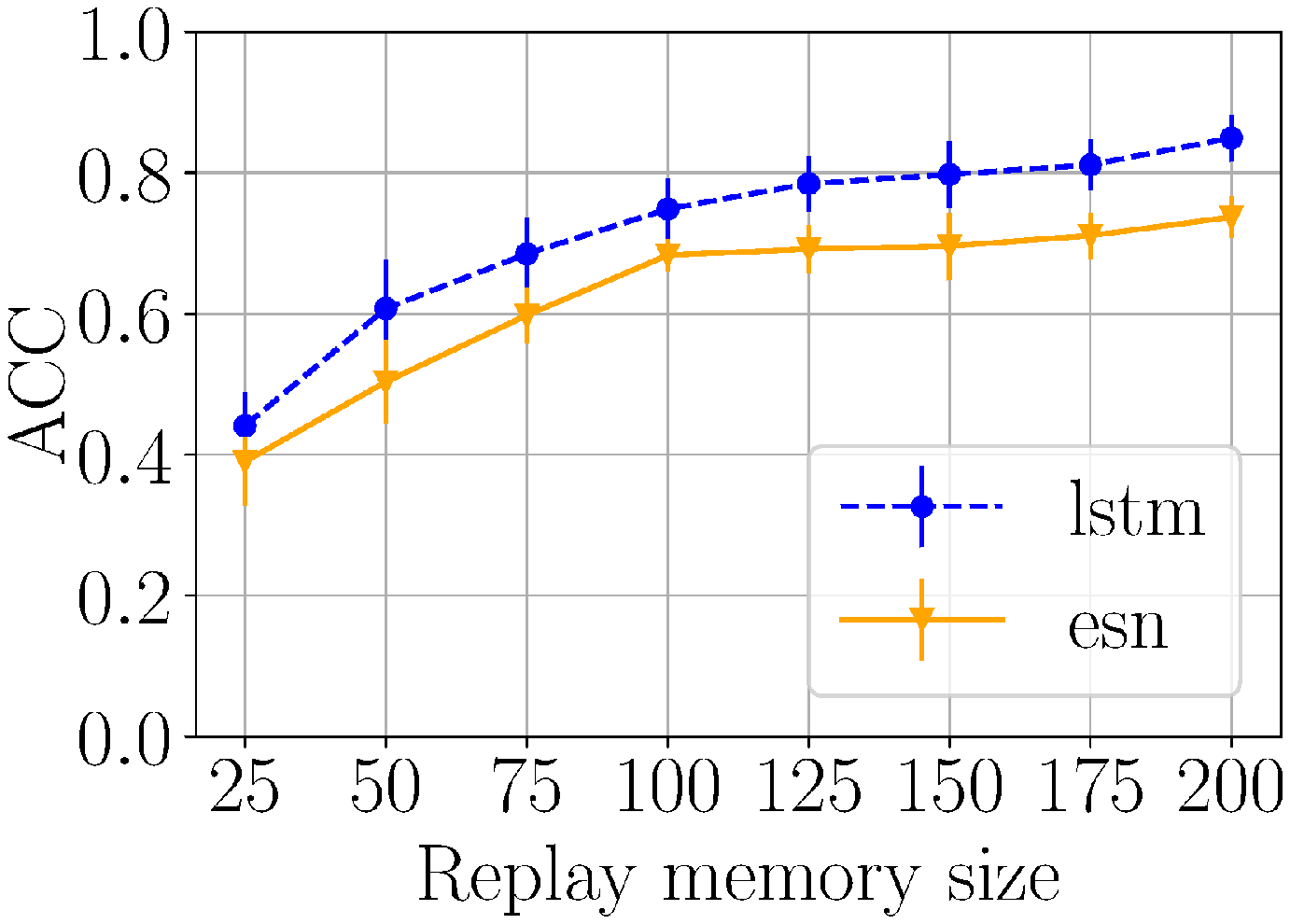}
    \caption{Split MNIST}
\end{subfigure}
\begin{subfigure}{.42\textwidth}
    \centering
    \includegraphics[width=\textwidth]{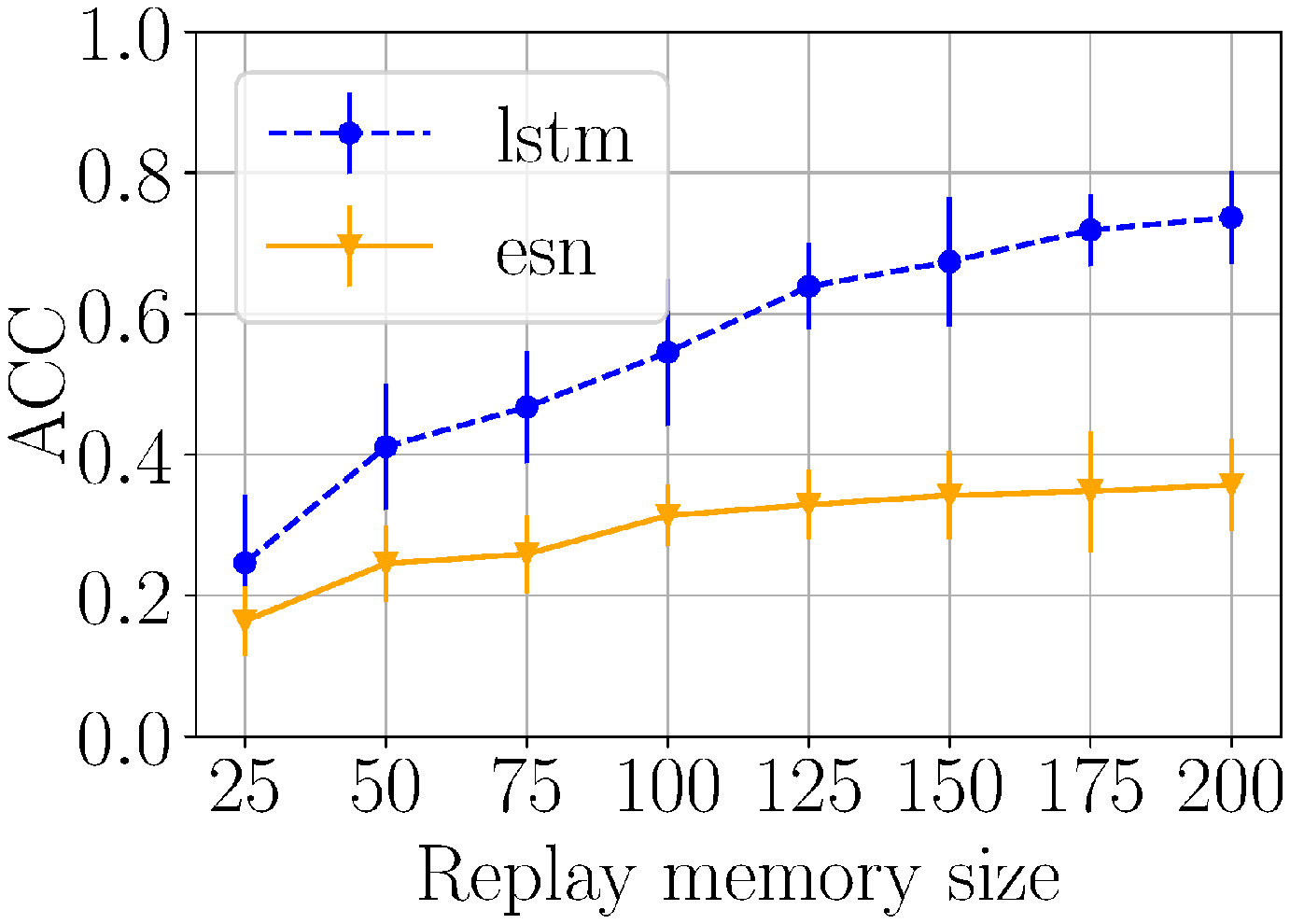}
    \caption{SSC}
    \end{subfigure}
\caption{Accuracy over increasing replay memory sizes.}
\label{fig:replay}
\end{figure}

Table \ref{tab:results} reports the ACC metric and its standard deviation over $5$ runs for the best configuration found in model selection. Our results show that ESN performs better or comparably to the LSTM network. EWC is not able to tackle class-incremental scenarios, neither with LSTM nor with ESN. It achieves a performance equal to the Naive one. LwF, instead, manages to reduce forgetting in the simplest SMNIST benchmarks. In this context, applying LwF only on the feedforward component results in a better performance with respect to the LSTM. This is compatible with results presented in \cite{cossu2021} and highlights one of the advantage in using ESNs for CL. However, when facing more complex scenarios like SSC, LwF fails to provide any benefits with respect to Naive finetuning. For replay, we studied the performance of ESN and LSTM with different memory sizes (Fig. \ref{fig:replay}). ESN exhibits a lower performance with respect to a fully trained LSTM. On SMNIST, the gap becomes significant only for large memory sizes. On SSC, instead, the difference in performance occurs also for smaller memories. This result is surprising since replay is not usually sensitive to the choice of the model. A deeper investigation of different replay policies and ESNs architectures will be needed in order to discover possible solutions to the problem. \\
The advantage in using ESNs clearly emerges when studying the behavior of SLDA strategy. This strategy effectively tackles class-incremental benchmarks like SSC. The absolute performance of SLDA in SSC is still far from the joint training upper bound. However, this does not mean that SLDA suffers from large forgetting effect. In fact, it achieves an average experience forgetting (difference between the accuracy after training on a certain experience and the corresponding accuracy after training on all experiences) of $0.14 \pm 0.02$. The remaining $20\%$ of difference from the joint training is explained by the fact that SLDA is a streaming strategy which trains the model only on a single epoch. Therefore, on complex scenarios like SSC it achieves a lower accuracy. 

\section{Conclusion and Future Work}
We studied the ability of ESNs to mitigate forgetting in CL environments. We provided the first experimental evaluation of ESNs trained with popular CL strategies. Our analysis showed that, apart from replay strategies, ESNs perform comparably with fully trained recurrent networks like LSTM. Moreover, since the reservoir is a fixed feature extractor, it is possible to train ESNs with CL strategies like SLDA which are not applicable to LSTM. SLDA obtains a good performance in class-incremental scenarios.\\
This work could foster new studies and applications of CL with ESNs: the renowned computational efficiency of ESNs may be particularly interesting for streaming or task-free CL. The possibility to implement ESNs in neuromorphic hardware opens to the continuous training of such models on low resources devices. ESNs are not the only family of models with an untrained component. More in general, CL with partially trained networks constitutes an interesting avenue of research, due to the fact that fixed connections are not subjected to catastrophic forgetting. Alternatively, reservoir in ESNs may also be finetuned during training to better adapt to new experiences. In particular, unsupervised finetuning through backpropagation-free methods (e.g. Hebbian learning, intrinsic plasticity) may provide quicker adaptation and more robust representations. The design of new CL strategies which exploit reservoir finetuning while keeping forgetting into consideration would provide us with a deeper understanding of the CL learning capabilities of ESNs.

\begin{footnotesize}

\bibliographystyle{unsrt}
\bibliography{CL.bib}

\begin{thebibliography}{10}

\bibitem{lesort2020}
Timoth{\'e}e Lesort, Vincenzo Lomonaco, Andrei Stoian, Davide Maltoni, David
  Filliat, and Natalia {D{\'i}az-Rodr{\'i}guez}.
\newblock Continual learning for robotics: {{Definition}}, framework, learning
  strategies, opportunities and challenges.
\newblock {\em Information Fusion}, 58, 2020.

\bibitem{Jaeger2004}
Herbert Jaeger and Harald Haas.
\newblock Harnessing nonlinearity: Predicting chaotic systems and saving energy
  in wireless communication.
\newblock {\em Science}, 304, 2004.

\bibitem{cossu2021}
Andrea Cossu, Antonio Carta, Vincenzo Lomonaco, and Davide Bacciu.
\newblock Continual {{Learning}} for {{Recurrent Neural Networks}}: An
  {{Empirical Evaluation}}.
\newblock {\em arXiv}, 2021.

\bibitem{sodhani2019}
Shagun Sodhani, Sarath Chandar, and Yoshua Bengio.
\newblock Toward {{Training Recurrent Neural Networks}} for {{Lifelong
  Learning}}.
\newblock {\em Neural Computation}, 32, 2019.

\bibitem{duncker2020}
Lea Duncker, Laura~N Driscoll, Krishna~V Shenoy, Maneesh Sahani, and David
  Sussillo.
\newblock Organizing recurrent network dynamics by task-computation to enable
  continual learning.
\newblock In {\em Advances in {{Neural Information Processing Systems}}},
  volume~33, 2020.

\bibitem{cossu2020}
Andrea Cossu, Antonio Carta, and Davide Bacciu.
\newblock Continual {{Learning}} with {{Gated Incremental Memories}} for
  sequential data processing.
\newblock In {\em International Joint Conference on Neural Networks}, 2020.

\bibitem{ororbiaSpikingNeuralPredictive2020}
Alexander Ororbia.
\newblock Spiking {{Neural Predictive Coding}} for {{Continual Learning}} from
  {{Data Streams}}.
\newblock {\em arXiv}, 2020.

\bibitem{kobayashi2019}
Taisuke Kobayashi and Toshiki Sugino.
\newblock Continual {{Learning Exploiting Structure}} of {{Fractal Reservoir
  Computing}}.
\newblock In {\em ICANN}, 2019.

\bibitem{kirkpatrick2017}
James Kirkpatrick, Razvan Pascanu, Neil Rabinowitz, Joel Veness, Guillaume
  Desjardins, Andrei~A Rusu, Kieran Milan, John Quan, Tiago Ramalho, Agnieszka
  {Grabska-Barwinska}, Demis Hassabis, Claudia Clopath, Dharshan Kumaran, and
  Raia Hadsell.
\newblock Overcoming catastrophic forgetting in neural networks.
\newblock {\em PNAS}, 114, 2017.

\bibitem{li2016}
Zhizhong Li and Derek Hoiem.
\newblock Learning without {{Forgetting}}.
\newblock In {\em European {{Conference}} on {{Computer Vision}}}, 2016.

\bibitem{robins1995}
Anthony Robins.
\newblock Catastrophic {{Forgetting}}; {{Catastrophic Interference}};
  {{Stability}}; {{Plasticity}}; {{Rehearsal}}.
\newblock {\em Connection Science}, 7, 1995.

\bibitem{prabhu2020}
Ameya Prabhu, Philip H.~S. Torr, and Puneet~K. Dokania.
\newblock {{GDumb}}: {{A Simple Approach}} that {{Questions Our Progress}} in
  {{Continual Learning}}.
\newblock In {\em {{ECCV}}}, 2020.

\bibitem{hayes2020}
Tyler~L Hayes and Christopher Kanan.
\newblock Lifelong {{Machine Learning}} with {{Deep Streaming Linear
  Discriminant Analysis}}.
\newblock In {\em {{CLVision Workshop}} at {{CVPR}}}, 2020.

\bibitem{lukoseviciusReservoirComputingApproaches2009}
Mantas Luko{\v s}evi{\v c}ius and Herbert Jaeger.
\newblock Reservoir computing approaches to recurrent neural network training.
\newblock {\em Computer Science Review}, 3, 2009.

\bibitem{lomonaco2021}
Vincenzo Lomonaco, Lorenzo Pellegrini, Andrea Cossu, Antonio Carta, Gabriele
  Graffieti, Tyler~L. Hayes, Matthias De~Lange, Marc Masana, Jary Pomponi, Gido
  {van de Ven}, Martin Mundt, Qi~She, Keiland Cooper, Jeremy Forest, Eden
  Belouadah, Simone Calderara, German~I. Parisi, Fabio Cuzzolin, Andreas
  Tolias, Simone Scardapane, Luca Antiga, Subutai Amhad, Adrian Popescu,
  Christopher Kanan, Joost {van de Weijer}, Tinne Tuytelaars, Davide Bacciu,
  and Davide Maltoni.
\newblock Avalanche: An {{End}}-to-{{End Library}} for {{Continual Learning}}.
\newblock In {\em {{CLVision Workshop}} at {{CVPR}}}, 2021.

\end{thebibliography}

\end{footnotesize}

\end{document}